\documentclass[10pt,twocolumn,letterpaper]{article}

\usepackage{cvpr}
\usepackage{times}
\usepackage{epsfig}
\usepackage{graphicx}
\usepackage{amsmath}
\usepackage{amssymb}
\usepackage{algorithm,graphicx,xcolor,tabularx}
\usepackage{algorithmic}
\usepackage{subfigure}
\usepackage{color}
\usepackage{booktabs}
\usepackage{threeparttable}
\usepackage{enumitem}
\usepackage{times}
\usepackage{helvet}
\usepackage{courier}
\usepackage{tabularx,url,array,nth,graphicx,subfigure}
\usepackage{booktabs}
\usepackage{threeparttable}
\usepackage{enumitem}
\usepackage{color,xcolor}
\usepackage[breaklinks=true,bookmarks=false]{hyperref}
\cvprfinalcopy 
\def\cvprPaperID{1633} 

\newcommand{\czq}[1]{\textcolor{black}{#1}}

\ifcvprfinal\fi
\begin{document}

\title{Video2Shop: Exact Matching Clothes in Videos to Online Shopping Images}
\author{	
Zhi-Qi Cheng\textsuperscript{1,2}, Xiao Wu\textsuperscript{1}, Yang Liu\textsuperscript{2}, Xian-Sheng Hua\textsuperscript{2}\\
\textsuperscript{1}Southwest Jiaotong University, \textsuperscript{2}Alibaba Group\\
{\tt\small \{zhiqicheng,huaxiansheng\}@gmail.com; wuxiaohk@swjtu.edu.cn; panjun.ly@alibaba-inc.com}
}
\maketitle
\thispagestyle{empty}


\begin{abstract}
	In recent years, both online retail and video hosting service are exponentially growing.
	In this paper, we explore a new cross-domain task, Video2Shop, targeting for matching clothes appeared in videos to the exact same items in online shops.
	A novel deep neural network, called AsymNet, is proposed to explore this problem.
	For the image side, well-established methods are used to detect and extract features for clothing patches with arbitrary sizes.
	For the video side, deep visual features are extracted from detected object regions in each frame, and further fed into a Long Short-Term Memory (LSTM) framework for sequence modeling, which captures the temporal dynamics in videos.
	To conduct exact matching between videos and online shopping images, LSTM hidden states, representing the video, and image features, which represent static object images, are jointly modeled under the similarity network with reconfigurable deep tree structure.
	Moreover, an approximate training method is proposed to achieve the efficiency when training.
	Extensive experiments conducted on a large cross-domain dataset have demonstrated the effectiveness and efficiency of the proposed AsymNet, which outperforms the state-of-the-art methods.
	
\end{abstract}

\section{Introduction}
Online retail is growing exponentially in recent years, among which the clothing shopping occupies a large proportion.
Driven by the huge profit potential, intelligent clothing item retrieval is receiving a great deal of attention in the multimedia and computer vision literature.
Meanwhile, online video streaming service is becoming increasingly popular. When watching idol drama or TV shows, such as the Korean TV drama \textit{My Love From the Star}, where beautiful girls wearing fashion clothes, the viewers are more easily attracted by those beautiful clothes and stimulated to buy the identical ones shown in the video, especially the females. In this paper, we consider a new scenario of such online clothing shopping: finding the clothes identical to the ones worn on the actors during watching videos. We call this new search approach as \textit{Video2Shop}.

\begin{figure*}[tb]
	\centering
	\includegraphics[scale=0.45]{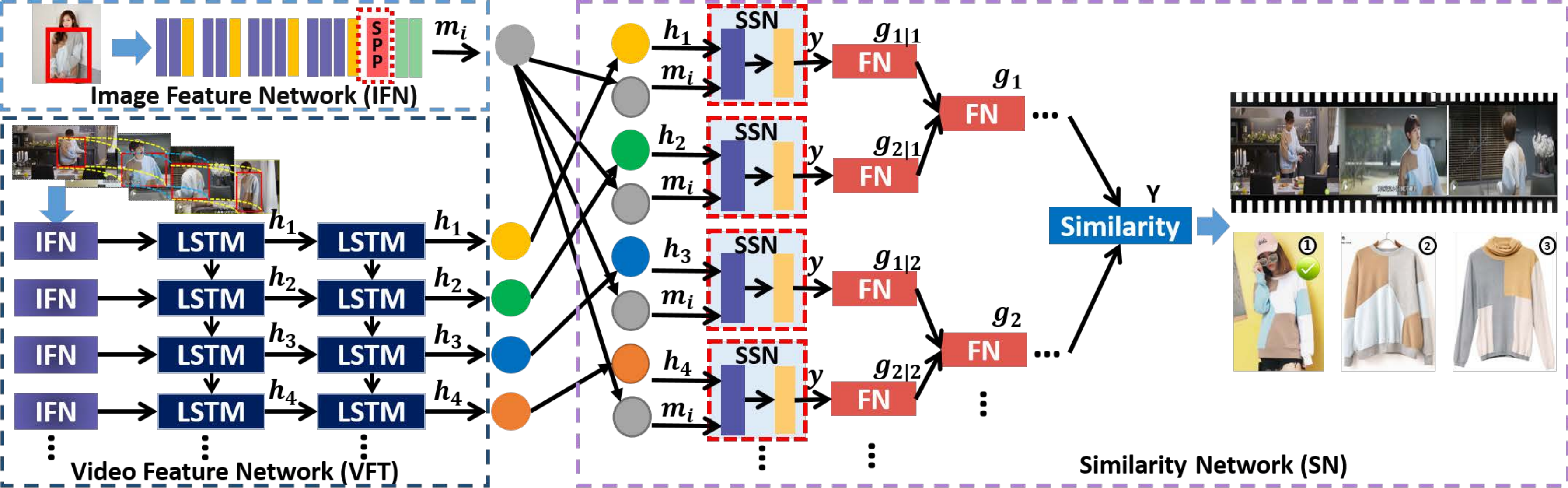}
	\caption{Framework of the proposed AsymNet. After clothing detection and tracking, deep visual features are generated by image feature network (IFN) and video feature network (VFN), respectively. These features are then fed into the similarity network to perform pair-wise matching.
	}
	\label{fig:FR}
	\vspace{-0.2in}
\end{figure*}

Although the street-to-shop clothing matching problem, which searches the online clothing by street fashion photos, has been explored recently \cite{ICCV15_CD, MM14_DS, ICCV15_Wheretobuy, ICMR16_Product}, finding clothes appeared in videos to the exact same items in online shops is not well studied yet. The diverse appearance of cloth, cluttered scenes, occlusion, different light condition and motion blur in the video make video2shop challenging.
More specifically, the clothing items appeared in videos and online shopping websites demonstrate significant visual discrepancy. On one hand, in the video, the clothes are usually captured from different viewpoints, (the front, the side or the back), or following the path of the actors, which leads to great varieties in clothes appearance. The complex scenes and the common motion blur in videos even make the situation worse. On the other hand, the online clothing images are not always with clean background, since the clothes are often worn by fashion models in outdoor scenes to show its real wearing effect. The cluttered background imposes difficulties for clothing localization and analysis. These problems caused by the videos and the online clothing images make the \textit{Vidoe2Shop} task more challenging than the street-to-shop search.

The architecture of the proposed a deep neural network, AsymNet, is illustrated in Fig. \ref{fig:FR}. When users watch videos through web pages or set-top box devices, the system will retrieve the exact matched clothing items from online shops and return them to the users. Clothing detector is first deployed for both video side and image side, to extract a set of proposals (clothing patches) to identify the potential clothing regions, limiting the impact of background regions and leading to more accurate clothing localization. For videos, clothing tracker is then conducted to track clothing patches to generate clothing trajectory, which contains the same clothing items appeared in continuous frames. Intuitively, the clothing patches with inconsistent viewpoints are preserved. Due to their promising performance and stability, Faster-RCNN \cite{NIPS15_FasterR-CNN} and Kernelized Correlation Filters (KCF) tracker \cite{TPAMI15_KCF} are adopted in this paper as the clothing detector and clothing tracker, respectively. Deep visual features are generated for clothing images in shops and clothing trajectories in videos, which are achieved with image feature network (IFN) and video feature network (VFN), respectively. For videos, deep visual features are further fed into a Long Short-Term Memory (LSTM) framework \cite{AR14_LSTM} for sequence modeling, which captures the temporal dynamics in videos. To consider the whole clothing trajectories, this problem is formulated as an asymmetric (multiple-to-single) matching problem, i.e., exact matching a sequence of a cloth appeared in videos to a single online shopping clothing. These features are then fed into the similarity network to perform pair-wise matching between clothing regions from videos and shopping images, in which a reconfigurable deep tree structure is proposed to automatically learn the fusion strategy. The top ranked results are then returned to users.

The main contributions of the proposed work are summarized as follows:
\begin{itemize}[leftmargin=*]
	\item A novel deep-based network, AsymNet, is proposed for cross-domain Video2Shop application, which is formulated as an asymmetric (multiple-to-single) matching problem. It mainly consists of two components: image and video feature representation and similarity measure.

	\item To conduct exact matching, LSTM hidden states for clothing trajectories in videos, and image features representing online shopping images, are jointly modeled under the similarity network with a reconfigurable deep tree structure.
	
	\item To train AsymNet, an approximate training method is proposed to improve the training efficiency. The proposed method can handle the large-scale online search.

	\item Experiments conducted on the first and the largest Video2Shop dataset demonstrate the effectiveness of the proposed method, which consists of 26,352 clothing trajectories in videos and 85,677 clothing images from shops. The proposed method outperforms the state-of-the-art approaches.
\end{itemize}	

The rest of our paper is organized as follows: related works are first reviewed in Section \ref{sec:rw}. The details of feature extraction networks and similarity networks are elaborated in Sections \ref{sec:Feature} and \ref{sec:SN}, respectively. The approximate training of the network is presented in Section \ref{sec:Train}. Finally, experiments are introduced in Section \ref{sec:ex}.

\section{Related Work}
\label{sec:rw}

\subsection{Cross-Scenario Clothing Retrieval}
Cross-scenario clothing retrieval has widely applicability for commercial systems.
There have been extensive efforts on similar clothing retrieval \cite{TOG15_CS,ICCV15_CD,MM14_DS,TMM16_CoPars,ICMR13_Recognitionandsegmentation,CVPR12_Street-to-shop}
and exactly same clothing retrieval \cite{ICCV15_Wheretobuy,ICMR16_Product}.

For similar clothing retrieval,
clothing recognition and segmentation techniques are used in \cite{TMM16_CoPars,ICMR13_Recognitionandsegmentation} to retrieve similar clothing. In order to tackle the domain discrepancy between street photos and shop photos, sparse representations are utilized in \cite{CVPR12_Street-to-shop}. With the adoption of deep learning,
an attribute-aware fashion-related retrieval system is proposed in \cite{MM14_DS}. A convolutional neural network using the contrastive loss is proposed in \cite{TOG15_CS}.
Based on the Siamese network, a Dual Attribute-aware Ranking Network (DARN) is proposed in \cite{ICCV15_CD}.

For exactly same clothing retrieval,
exact matching street clothing photos in online shops
is firstly explored in \cite{ICCV15_Wheretobuy}.
A robust deep feature representation is learned in \cite{ICMR16_Product} to bridge the domain gap between the street and shops.
A new deep model, namely FashionNet, is proposed in \cite{CVPR16_DeepFashion}, which learns clothing features by jointly predicting clothing attributes and land-marks.
Despite recent advances in exactly street-to-shop retrieval, there have been rather few studies focused specifically on exact matching clothes in videos to online shops.

\subsection{Deep Similarity Learning}

As deep convolutional neural networks are becoming ubiquitous, there has been growing interest in similarity learning with deep models.
For image patch-matching, some convolutional neural networks are proposed in \cite{CVPR15_MatchNet,CVPR15_SIM,CVPR15_CSIM_lecun}.
These techniques learn representations coupled with either pre-defined distance functions, or with more generic learned multi-layer network similarity measures.
For object retrieval, an neural network with contrastive loss function is designed in \cite{TOG15_CS}.
A novel \czq{Deep Fashing Network architecture is proposed in \cite{CVPR16_DeepFashion}} for efficient similarity retrieval.
Inspired by these works, we propose a tree structure similarity learning networks to match clothes appeared in videos to the exact same items in online shops.

\section{Representation Learning Networks}
\label{sec:Feature}

When the clothing regions are detected in images and then tracked into clothing trajectories for videos, feature extraction networks are then conducted to obtain the deep features.

\subsection{Image Representation Learning Networks}
\label{sec:IFN}
\begin{figure}[tb]
	\centering
	\includegraphics[scale=0.6]{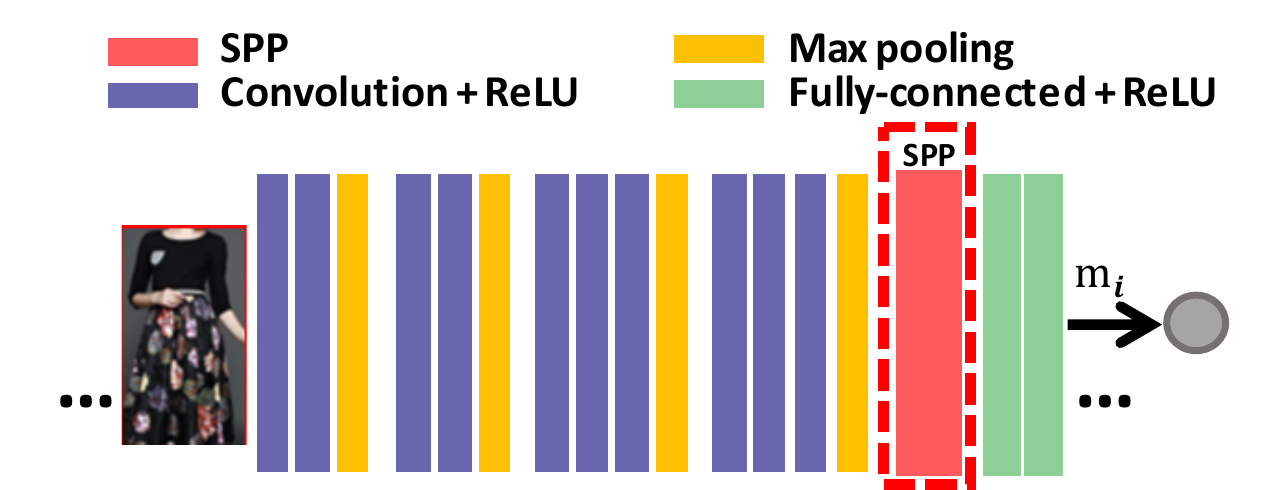}
	\caption{The Architecture of Image Feature Network}
	\label{fig:N1}
	\vspace{-0.2in}
\end{figure}

The image feature network (IFN) is implemented based on VGG16 \cite{AR16_VGG16}.
In VGG16, the input image patches are scaled to 256x256 and then cropped to a random 227x227 region.
This requirement comes from the fact that the output of the last convolutional layer of the network needs to have a predefined dimension.
In our Video2Shop matching task, Faster-RCNN \cite{NIPS15_FasterR-CNN} is adopted to detect clothing regions in the shopping images. Unfortunately, the detected clothing regions are with arbitrary sizes, which does not meet the requirement of the input size. Enlightened by the idea of the recently proposed spatial pyramid pooling (SPP) architecture \cite{TPAMI_SPP}, which pool features in arbitrary regions to generated fixed-length representations, a spatial pyramid pooling layer is inserted between the convolutional layers and the fully-connected layers of the network in VGG16, as shown in Fig. \ref{fig:N1}. It aggregates features of the last convolutional layer through spatial pooling, so that the size of the pooling regions is independent of the size of the input.

\subsection{Video Representation Learning Networks}
\label{sec:VFN}
Video Feature Network (VFN) is illustrated in Fig. \ref{fig:FR}.
For videos, the aforementioned image feature network (IFN) is also used to extract convolutional features.
Since the temporal dynamics exist in videos, traditional average pooling strategy becomes invalid.
Recurrent neural network is a perfect choice to solve this problem.
Recently, due to its long short-term memory capability for modeling sequential data, Long Short-Term Memory (LSTM) \cite{AR14_LSTM} has been successfully applied to a variety of sequence modeling tasks.
In this paper, it is chosen to characterize the clothing trajectories in videos.

Based on the LSTM unit proposed in \cite{AR14_Lstm_cell}, a typical LSTM unit consists of an input gate $i_t$, a forget gate $f_t$, an output gate $o_t$, as well as a candidate cell state $g_t$. The interaction between states and gates along the time dimension is defined as follows:
\begin{align}
\begin{pmatrix}
\mathbf{i}_t\\ \mathbf{f}_t\\ \mathbf{o}_t\\ \mathbf{g}_t
\end{pmatrix} &=
\begin{pmatrix}
\sigma \\ \sigma \\ \sigma \\ \text{tanh}
\end{pmatrix}
M
\begin{pmatrix}
\mathbf{h}_{t-1}\\ \mathbf{m}_t
\end{pmatrix},\nonumber \\
\mathbf{c}_t &= \mathbf{f}_t \odot \mathbf{c}_{t-1} + \mathbf{i}_t \odot \mathbf{g}_t,\\
\mathbf{h}_t &= \mathbf{o}_t \odot \tanh\left(\mathbf{c}_t\right). \nonumber
\end{align}
Here, $c_t$ encodes the cell state, $h_t$ encodes the hidden state, and $m_t$ is the convolutional feature generated by the image feature network. The operator $\odot$ represents element-wise multiplication.
Given convolutional features $M(m_1 ,...,m_n)$ of a clothing trajectory in videos, a single LSTM computes a sequence of hidden states $(h_1 ,...,h_n )$.
Further, we find that the temporal variety cannot be fully learned by a single LSTM, so we stack LSTM network to further increase the discriminative ability of the network, by using the hidden units from one layer as inputs for the next layer.
After experimental validation, a two-level LSTM network is utilized in this work.

\section{Similarity Learning Networks}
\label{sec:SN}

\subsection{Motivation}

To conduct pair-wise similarity measure between clothing trajectories from videos and shopping images, a similarity network is proposed. The inputs are several LSTM hidden states $(h_1, h_2,..., h_n)$ from video feature network and a convolutional feature $m_i$ from image feature network. The output is a similarity score $Y$.
This problem is formulated as an asymmetric (multiple-to-single) matching problem. Traditionally, this problem is solved by conducting average or max pooling on whole clothing trajectories to obtain the global similarity or directly select the similarity of the last one in trajectories. More recently, a key volume detection method \cite{CVPR2016_Keyvolume} is also proposed to solve the similar problem. However, these methods will fail in our Video2Shop application due to the large variability and complexity of video data. The average or max values cannot completely represent the clothing trajectory. Although key volume is able to learn the most critical parts, it is still too simple to solve this task.

Based on the statistical theory \cite{NC91_ME1,NC94_ME2}, these learning problems are formulated as a mixture estimation problem, which attacks a complex problem by dividing it into simpler problems whose solutions can be combined to yield a solution to the complex problem. Enlightened by this idea, we novelly extend the generalized mixture expert model to Recurrent Neural Networks (RNN), and modify the strategy of mixture estimation to gain a global similarity. The proposed approach attempts to allocate fusion nodes to summarize the single similarity located in different viewpoints.

\subsection{Network Structure}

Because there are multiple inputs and only one output, a tree structure is proposed to automatically adjust the fusion strategy, which is illustrated in Fig. \ref{fig:FR}.
There are two types nodes involved in the tree structure, i.e., single similarity network node (SSN) and fusion nodes (FN), corresponding to the leaves and the branches in the tree.
The single similarity network (SSN) acts as the leaves of the tree, which calculates the similarity between a single LSTM hidden state $h_i$ and a convolutional feature $m_i$. After that, these results are passed to Fusion Node (FN), which generates a scalar output controlling the weights of similarity fusion. These fusion nodes will be passed layer by layer to fuse the results of internal results. In this work, a five-layer structure is adopted. Finally, a final global similarity $Y$ will be given.
Details of each substructure are given below.

\paragraph{Single Similarity Network (SSN)}
To facilitate understanding, we will first introduce the one-to-one similarity measure between a LSTM hidden state $h_i$ and a convolutional feature \czq{$m_i$}. As indicated in \cite{ICCV15_Wheretobuy}, cosine similarity is too general to capture the underlying differences between features.
Therefore, the similarity between $h_i$ and \czq{$m_i$} is modeled as a network with two fully-connected layers, denoted as the red dotted box shown in Fig. \ref{fig:FR}.
Specifically, the first two fully-connected layers have 256 (fc1) and 1 (fc2) outputs, respectively.
The output of the last fully-connected layer is a real value $z$. On the top of the network, logistic regression is used to
generated the similarity between $h_i$ and \czq{$m_i$} as:
\begin{equation}
\label{eqn_1}
\hat{y}=\frac{1}{1+e^{-z}}
\end{equation}
\paragraph{Fusion Node (FN)}
Since SSN is piece-wisely smoothed, which is analogous to corresponding generalized linear models (GLIM) \cite{ML1987_GLIM}.
Once the individual SSN is calculated, the fusion nodes (FN) at lower levels will integrate the results of SSN and control their weights, which are defined as a generalized linear system \cite{ML1994_HME}.
The intermediate variable $\mathbf{\varepsilon_{ij}}$ is defined as:
\begin{equation}
\label{eqn_Fn_1}
\mathbf{\varepsilon_{ij}}=\mathbf{v_{ij}}^{T}\left (\mathbf{x_{ij}} \right )
\end{equation}
where \czq{ 
	the subscript $i$ and $j$ denotes the index of fusion nodes, in which $i$ and $j$ refer to the low-level and high-level FN nodes, as Fig. \ref{fig:FR}}, $\mathbf{\mathbf{v_{ij}}}$ is a weight vector, $\mathbf{x_{ij}}$ is a feature \czq{vector} of the fc1 layer.
\czq{The output of lower levels of the fusion node is a product of $g_{i|j}$ (output of
Eqn. \ref{eqn_Fn_2}) times $\hat{y}$ (output of SSN). The $g_{i|j}$ is a scale as}:
\begin{equation}
\label{eqn_Fn_2}
g_{i|j}= \frac{e^{\varepsilon_{i,j}}}{\sum_{k}e^{\varepsilon_{i,j}}}
\end{equation}
Note that, $g_{i|j}$ is positive and their sum is equal to one, which can be also interpreted as providing a local fusion for each top level fusion node.

Considering the hierarchical fusion strategy can obtain a better performance \cite{ML1994_HME}, the fusion nodes are constructed as a tree structure.
Similarly, the intermediate variable $\varepsilon_{j}$ is defined, and the weight vector $\mathbf{v_{j}}$ is defined as Eqn. \ref{eqn_Fn_1}.
In particular, \czq{$\mathbf{x_{j}}$ is an average pooling vector from multiple $\mathbf{x_{ij}}$}. 
The output $g_j$ of the top fusion node is also defined as Eqn. \ref{eqn_Fn_2}.
$g_j$ is positive and their sum is equal to one, which can be interpreted as providing a global fusion function.
With such a tree structure, for each mini-batch, we update the weights of fusion nodes in the forward pass. Once the similarity network converges, the global similarity is obtained.

\subsection{Learning Algorithm}
In this subsection, we will introduce the learning method of our similarity network.
The learning is implemented in a two-step iteration approach, where single similar network and fusion nodes will be mutually enhanced. The feature representation network and SSN are first learnt, and then the fusion nodes are learnt when SSN is fixed.

\paragraph{Learning of Single Similarity Network.}
The learning problem of SSN is defined as minimizing a Logarithmic Loss.
Suppose that we have $N$ convolutional features from the first fully-connected layer fc1 as $X = \{x_1 ,x_2 ...x_N \}$ and each has a label $y_i \in \{0,1\}$, where 0 means ``does not match'' while 1 means ``matches''. It is defined as:
\begin{equation}
\label{eqn:SN1}
L(W)=\frac{1}{N}\sum_{i=1}^{n}(y_i log\left ( \hat{y_i} \right ) + \left ( 1-y_i \right ) log\left ( 1-\hat{y_i} \right ))+\lambda \left \|W  \right \|^{2}
\end{equation}
where $W$ is the parameters of SSN, $y_i = 1$ for positive examples and $y_i = 0$ for negative examples. $\hat{y_i}$ is the output of single similarity network.

\paragraph{Learning of Fusion nodes.}
When SSN is fixed, for a given mini-batch feature set $X = \{x_1 ,x_2 ...x_N \}$ of the fc1 layer, the global similarity $Y$ can be defined as the mixture of the probabilities of generating $y_{ij}$ from each SSN:
\begin{equation}
\label{eqn:obj}
P(Y|X,\theta) = \sum_{j}g_{j}(X,\mathbf{v_{j}})\sum_{i}g_{j|i}(X,\mathbf{v_{ij}})p(y_{ij}|X,W_{ij})
\end{equation}
where $P(Y|X,\theta)$ and $p(y_{ij}|X,W_{ij})$ are global and single similarity.
$g_{j}(X,\mathbf{v_{j}})$ and $g_{j|i}(X,\mathbf{v_{ij}})$ are the weights of top and lower fusion nodes.
$\theta$ contains $\mathbf{v_{j}}$, $\mathbf{v_{ij}}$ and $W_{i,j}$, which are the weights of top fusion nodes, lower fusion nodes and SSN, respectively.

In order to implement the learning algorithms of Eqn. \ref{eqn:obj}, posterior probabilities of fusion nodes are defined.
The probabilities $g_j$ and $g_{j|i}$ are referred as prior probabilities, because they are computed based only on the input $\mathbf{x_{i}}$ from fc1 layer as Eqn. \ref{eqn_Fn_2}, without the knowledge of corresponding target output $y$ as described in SSN.
With Bayes' rule, the posterior probabilities at the nodes of the tree are denoted as follows:
\begin{equation}
\label{eqn:h1}
h_{j} = \frac{g_{j}\sum_{i}g_{j|i}P_{ij}(y)}{\sum_{j}g_{i}\sum_{i}g_{j|i}P_{ij}(y)}
\end{equation}
and
\begin{equation}
\label{eqn:h2}
h_{ij} = \frac{g_{j|i}P_{ij}(y)}{\sum_{j}g_{j|i}P_{ij}(y)}
\end{equation}

With these posterior probabilities, a gradient descent learning algorithm is developed for Eqn. \ref{eqn:obj}.
The log likelihood of a mini-batch dataset $X =\{x^{t},y^{t}\}_{1}^{N}$ is obtained:
\begin{equation}
l(\theta ;X)= \sum_{t}\ln \sum_{j}g_{i}^{(t)}\sum_{i}g_{j|i}^{(t)}P_{ij}(y^{(t)})
\end{equation}
In this case, by differentiating $l(\theta ;X)$ with respect to the parameters, the following gradient descent learning rules for the weight matrix are obtained.
\begin{equation}
\label{eqn:SN_2}
\bigtriangledown \mathbf{v_{j}}= \alpha \sum_{t}(h_{j}^{(t)}-g_{j}^{(t)})
\mathbf{x_{j}}^{(t)}
\end{equation}
\begin{equation}
\label{eqn:SN_3}
\bigtriangledown \mathbf{v_{ij}}= \alpha \sum_{t}h_{i}^{(t)}(h_{j|i}^{(t)}-g_{j|i}^{(t)})
\mathbf{x_{ij}}^{(t)}
\end{equation}
where $\alpha$ is a learning rate. These equations denote a batch learning algorithm to train fusion nodes (i.e. tree structure).
To form a deeper tree, each SSN is expanded recursively into a fusion node and a set of sub-SSN networks.
In our experiment, we have five-level deep tree structure and the number of fusion nodes in each level is 32, 16, 8, 4, 2, respectively.

\floatname{algorithm}{Algorithm}
\renewcommand{\algorithmicrequire}{\textbf{Input:}}
\renewcommand{\algorithmicensure}{\textbf{Output:}}
\begin{algorithm}[t]
	\caption{Approximate Training Method.}
	\label{alg:2}
	\begin{algorithmic}[1]
		\REQUIRE An AsymNet containing IFN, VFN and SSN, L: LSTM hidden states, C: convolutional feature.
		\ENSURE AsymNet
		\STATE Sample $n$ clothing trajectories and each trajectory $u$ has $2\times S$ clothing images;
		\STATE L= net\_foward(VFN), C= net\_foward(IFN);
		\STATE \czq{Copy L to $2\times S$ times as $\hat{L}$, sent C and $\hat{L}$ to SSN};
		\STATE Train SSN as Eqn. \ref{eqn:SN1} and compute $\bigtriangledown (SSN)$;
		\STATE Net\_foward(SSN) and compute $h_i$ and $h_{ij}$ as Eqn. \ref{eqn:h1}-\ref{eqn:h2}
		\STATE Train fusion nodes as Eqn. \ref{eqn:SN_2}-\ref{eqn:SN_3};
		\STATE Net\_backward(IFN; $\bigtriangledown (SSN)$);
		\STATE Net\_backward(VFT; $\bigtriangledown(VFN_u)$) as Eqn. \ref{eqn:vfn};	
	\end{algorithmic}
\end{algorithm}

\section{Approximate Training}
\label{sec:Train}
Intuitively, to achieve good performance, different models should be trained independently for different clothing categories.
To achieve this goal, a general AsymNet is first trained, followed by fine-tuning for each clothing category to achieve category specific models. \czq{There are 14 models to be
trained.} In this section, we will introduce the approximate training of AsymNet.

To train a robust model, millions of training samples are usually needed.
It is extremely time-consuming to train the AsymNet using traditional training strategy.
Based on an intrinsic property of this application,
that is, many positive and negative samples (i.e. shopping clothes) share the same clothing trajectories in the training stage,
an efficient training method is proposed, which is summarized in Alg. \ref{alg:2}.

Suppose that the batch size of training is $n$, so $n$ trajectories in videos are sampled. Meanwhile, for \czq{a single} trajectory $u$, $2\times S$ shopping images are sampled (the number of positives and negatives is equal to $S$). In total, we have $n$ clothing trajectories in videos and $2\times S \times n$ clothing images in shops in each batch. To achieve the acceleration of training, the LSTM hidden states of $n$ trajectories are copied $2\times S$ times and sent them to the similarity network to accelerate the similarity network training. In backward time, the gradient of each clothing trajectory can be approximated as
\begin{equation}
\label{eqn:vfn}
\bigtriangledown (VFN_u)= \frac{1}{2 \times S}\bigtriangledown (SSN_u)
\end{equation}
and gradient of clothing image in shops can be backward directly.

\section{Experiment}
\label{sec:ex}
In this section, we will evaluate the performance of individual component of AsymNet, and compare the proposed method with state-of-the-art approaches.

\subsection{Dataset and Metrics}
Without proper datasets available for Video2Shop application, we collect a new dataset to evaluate the performance of identical clothing retrieval through videos, which will be released later. To the best of our knowledge, this is the first and the largest dataset for Video2Shop application. There are a number of online stores in e-commerce websites Tmall.com and Taobao.com, which sell the same styles of clothes appeared in movies, TV and variety shows. Accordingly, the videos and corresponding online clothing images are also posted on these stores. We download these videos from Tmall MagicBox, a set-top-box device from Alibaba Group, and the frames containing the corresponding clothing are extracted as the clothing trajectories manually.
In total, there are 85,677 online clothing shopping images from 14 categories, 26,352 clothing trajectories are extracted from 526 videos through Tmall MagicBox, 39,479 exact matching pairs are obtained.  
We also collect similar matching pairs for evaluation of similar retrieval algorithms. The dataset information is listed in Table \ref{tab:performance}.

In order to train the clothing detector, 14 categories of clothes are manually labeled, in which 2000 positive samples are collected per category from online images.
Faster-RCNN \cite{NIPS15_FasterR-CNN} is utilized as the clothing detector, and the clothing trajectories are generated by Kernelized Correlation Filters (KCF) tracker \cite{TPAMI15_KCF}.
\czq{The parameters used in Faster RCNN and KCF are the same as the original version.}
Duplicate clothing trajectories are removed.
The length of the clothing trajectories is roughly equal to 32.
To maintain the temporal characteristics of clothing trajectories, a sliding window is used to unify the length of clothing trajectories into 32. Each clothing trajectory in our dataset is linked to exact matched clothing images and they are manually verified by annotators, which form the ground truth. With an approximate ratio of 4:1, these exact matching video-to-shop pairs are split into two disjoint sets (training and testing sets), which are nonoverlapped.
Meanwhile, in order to reduce the impact of background and lead to more accurate clothing localization.
Faster-RCNN is also used to extract a set of clothing proposals for online shopping images.

\textbf{Evaluation Measure:}
Since the category is assumed to be known in advance, the experiments are performed within the category.
Followed by the evaluation criterion of \cite{ICCV15_Wheretobuy,ICMR16_Product}, the retrieval performance is evaluated based on \emph{top-k accuracy}, which is the ratio of correct matches within the top k returned results to the total number of search. Notice that once there is at least one exactly same product among the top 5 results as the query, which is regarded as a correct match in our setup. For simplicity, the weighted average is used for evaluation.

\begin{figure}[tb]
	\centering
	\includegraphics[scale=0.23]{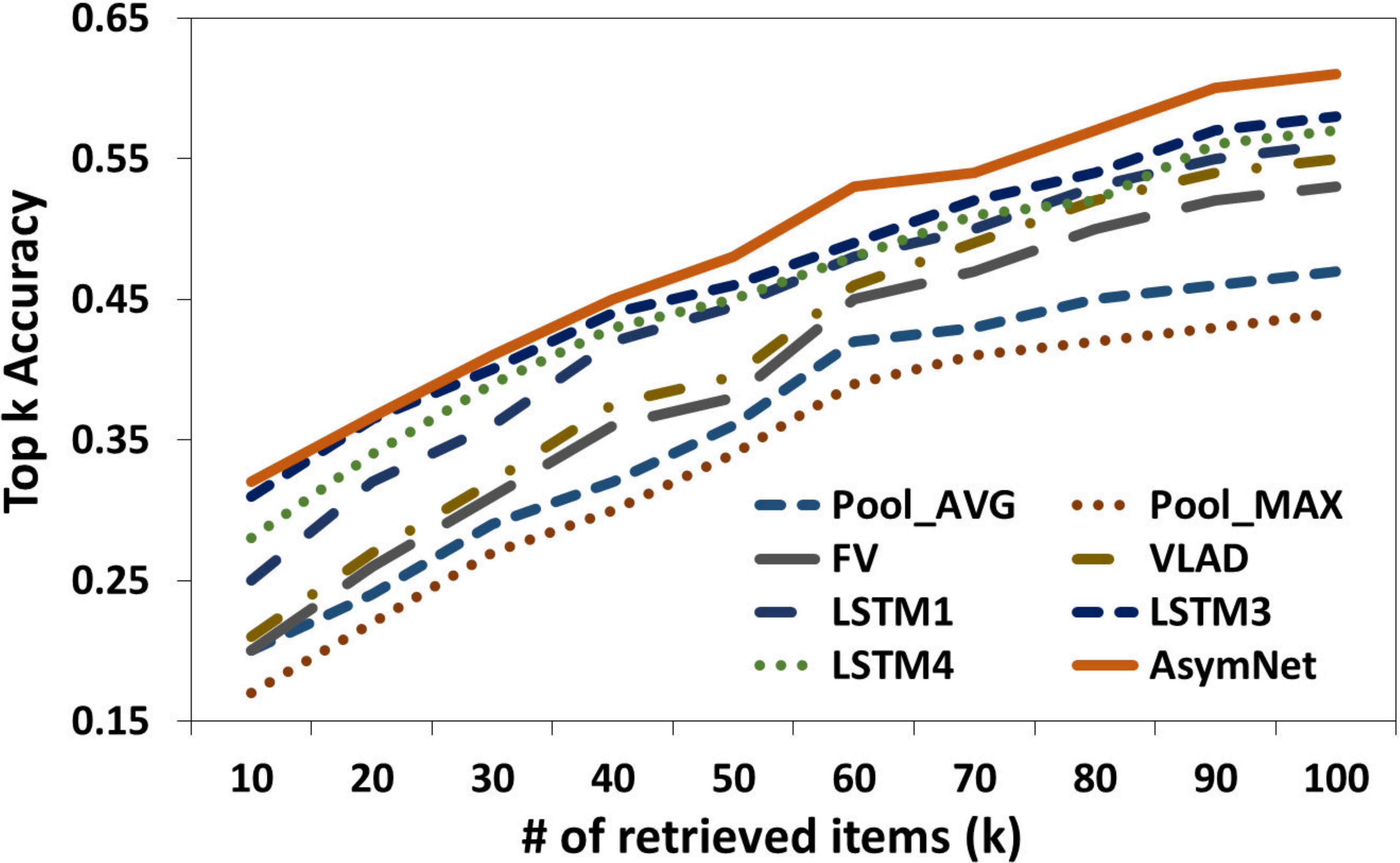}
	\caption{Performance Comparison of Representation Networks}
	\label{fig:F2}
	\vspace{-0.2in}
\end{figure}

\subsection{Performance of Representation Networks}
In this subsection, we compare the performance of representation networks with other baselines.
1) Average pooling,
2) Max pooling,
3) Fisher Vector \cite{CVPR07_Fishvector}
and 4) VLAD \cite{CVPR10_Vlad}.
We utilize 256 components for Fisher vectors and 256 centers for VLAD as common choices in \cite{CVPR10_Vlad,IJCV13_Fishvector}.
The PCA projections, GMM components of Fisher vectors, and K-means centers of VLAD are learned from approximately
18,000 sampled clothing regions in the training set.
For these baselines, average pooling and max pooling are directly used on the CNN features of clothing trajectories.
Fisher vector and VLAD are used to encode the CNN features of shopping images and clothing trajectories, respectively.
The similarity is then estimated by single similarity network.
In addition, the impact of different levels (1, 3 and 4 levels) of LSTM network is also investigated, denoted as LSTM1, LSTM3 and LSTM4, respectively. 
\czq{For LSTM based networks,
the final output from the similarity feature network is used
as the final matching result.} The performance comparison is shown in Fig. \ref{fig:F2}.

From Fig. \ref{fig:F2}, we can see that the general performance is increased as $k$ becomes larger, which means that it will be treated as a correct match once there is at least one exactly same item with the top $k$ returned results. But we can also noticed that the performance of top 10 is still far from satisfactory, since it still a challenging task to match clothes appeared in videos to the online shopping images. There exists significant discrepancy between these cross-domain sources, including diverse visual appearance, cluttered background, occlusion, different light condition, motion blur in the video, and so on.

The performance of average pooling is better than max pooling.
Both Fisher Vector and VLAD have better performance than the average pooling representation. And VLAD has slightly better performance than Fisher Vector.
Overall, all LSTM based networks outperform pooling based methods. The proposed AsymNet achieves the best performance, which has significantly higher performance than the other two pooling approaches.
As the increase of the levels of LSTM network, the performance is firstly increased and then dropped when the number of levels is more than two. Our AsymNet adopts the two-level LSTM structure.

\subsection{Structure Selection of Similarity Networks}

\begin{figure}[tb]
	\centering
	\includegraphics[scale=0.19]{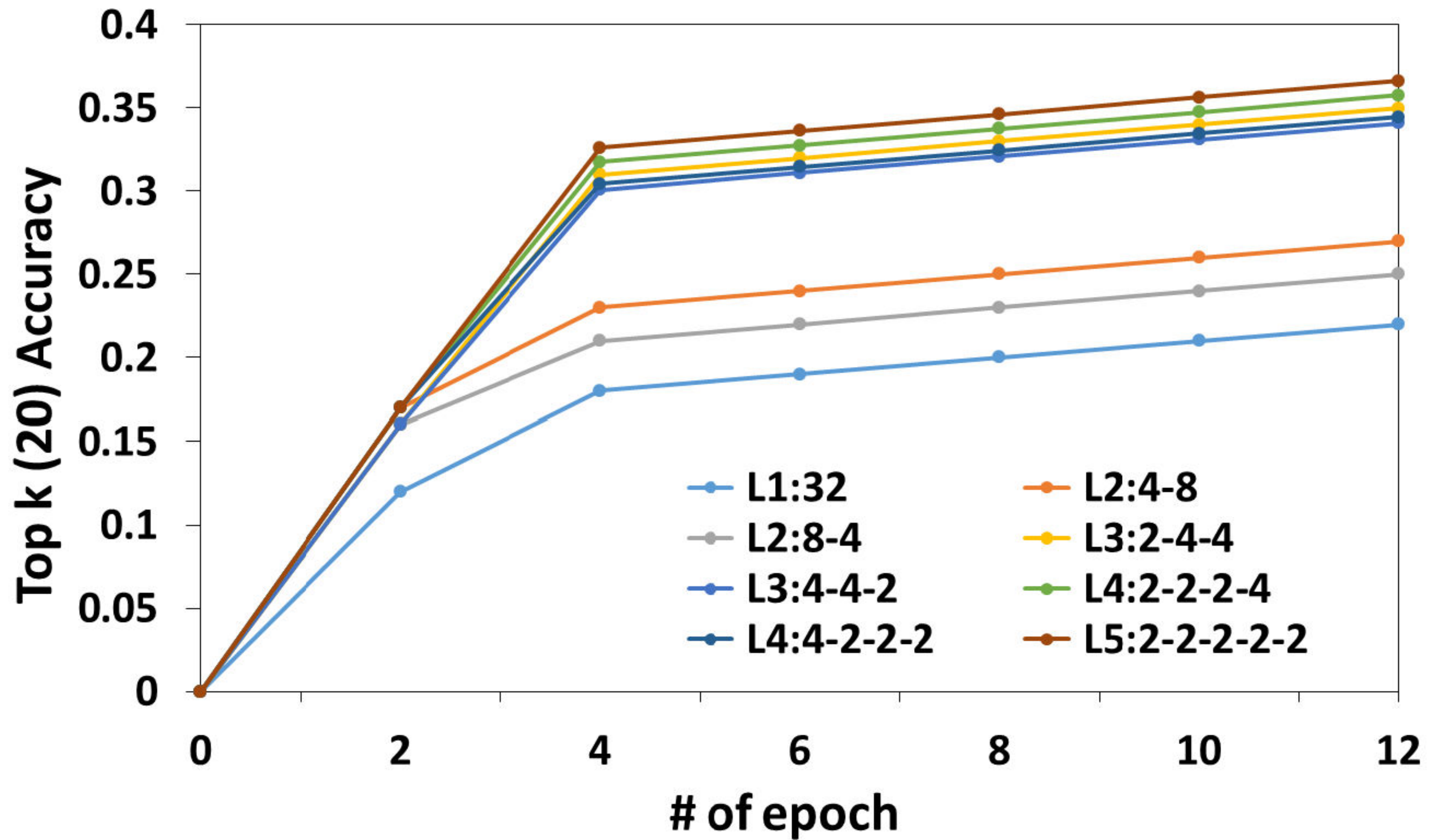}
	\caption{The top-20 retrieval accuracy (\%) of the proposed AsymNet with different structures.}
	\label{fig:F3}
	\vspace{-0.2in}
\end{figure}
To investigate the structure of similarity network, we vary the number of levels and the fusion nodes in similarity network, while keeping all other common settings fixed.
We evaluate two types of architectures:
1) Homogeneous branches: all fusion nodes have the same number of branches;
2) Varying branches: the number of branches is inconsistent across layers.
For homogeneous setting, one-level flat structure with 32 fusion nodes to hierarchical structure with five levels (62 fusion nodes) are tested. For the varying temporal branches, we compare six networks with branches in increasing order: 4-8, 2-4-4, 2-2-2-4 and decreasing order: 8-4, 4-4-2, 4-2-2-2, respectively.

The performance of these architectures is shown in Fig. \ref{fig:F3}, in which the structure is represented in the form: \#Level:\#Branches in each level from leaves to root of the tree, connected with hyphen. From this figure, we can see that the overall performance is significantly improved as the number of epoch increases. As the training proceeds,
the parameters in the fusion nodes begin to grow in magnitude, which means that the weights of fusion nodes are becoming more and more reasonable. Meanwhile, the performance is significantly improved as the number of epoch increases.
However, the improvement is not obvious after 4 epochs, since the weights of fusion nodes tends to be stable. The weight adjustment becomes subtle because the overall weights are optimized.

When one-level flat structure is adopted, it only has the leaves in the tree structure. The entire similarity network is reduced to a single averaged generalized linear models at the root of the tree.
As the training proceeds, the parameters in the fusion nodes begin to grow in magnitude.
When the fusion notes begin to take action, the performance of the system is boosted. We also notice that the general performance is increased when more levels of fusion nodes are involved. The boosting is pretty conspicuous for the first three layers. The improvement becomes minor when multi-level structure is formed. It indicates that the similar network becomes stable when the levels of fusion nodes are more than three.

\subsection{Performance of Similarity Learning Networks}
In order to verify the effectiveness of our similarity network, we compare the performance of the proposed method with other methods when fusion nodes are not included. These baselines include: the final matching result is determined by the average (Avg) and the maximum (Max) of all single similar networks, or the last (Last) single similar network. In addition, the latest work KVM \cite{CVPR2016_Keyvolume} is also considered, in which the key volume proposal method used in KVM is directly utilized to fuse the fc1 features in SSN. We formulate the similarity learning task as a binary classification problem.	With that, the same loss function in KVM can still be used.

\begin{figure}[tb]
	\centering
	\includegraphics[scale=0.23]{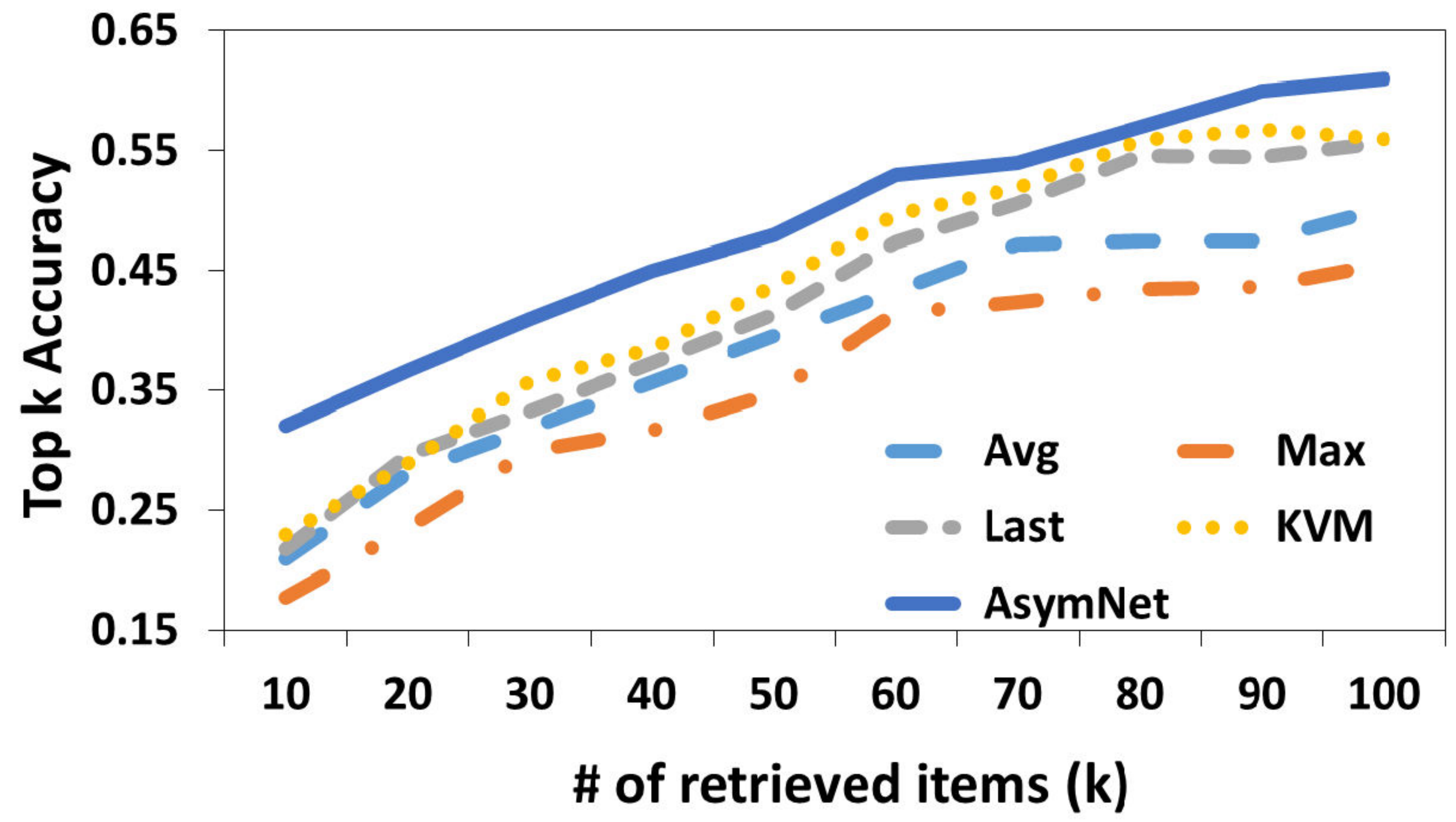}
	\caption{Performance of Similarity Learning Network}
	\label{fig:S1}
	\vspace{-0.2in}
\end{figure}

The top-20 retrieval performance comparison is shown in Fig. \ref{fig:S1}. From this figure, we can see that the performance of Avg is better than Max. Last has better performance than Avg and Max. The main reason is that the last hidden states learn the whole temporal information in the clothing trajectories. The noise in clothing trajectories affects the performance of Avg and Max greatly.
KVM considers the discriminative information may occur sparsity in a few key volumes, while other volumes are irrelevant to the final result. Although KVM is able to learn the most critical parts from clothing trajectories, it is too simple to consider the whole trajectory, in which different local viewpoints in trajectory is not well considered.
The proposed AsymNet outperforms these baselines, which has significantly higher performance. 

\begin{table*}[!t]
	\centering
	\caption{The top-20 retrieval accuracy (\%) of the proposed AsymNet compared with state-of-the-art approaches. The notations represent the numbers of images (\# I), video trajectories (\# TJ), queries (\# Q) and its corresponding results (\# R).}
	\label{tab:performance}
	\begin{tabularx}{17.5cm}{Xcccccccccc}
		\hline\hline
		\textbf{Category} &\textbf{\# I} &\textbf{\# TJ}   &\textbf{\# Q} & \textbf{\# R}  &\textbf{AL \cite{NIPS12_AlexNet}} & \textbf{DS \cite{MM14_DS}}  & \textbf{FT \cite{ICCV15_Wheretobuy}} &\textbf{CS \cite{TOG15_CS}}  &\textbf{RC \cite{ICMR16_Product}} &\textbf{AsymNet} \\
		\hline
		Outwear  & 18,144   & 5,581    & 1,116  & 3,628 &   17.31 & 22.94 & 26.97 & 27.61 & 31.80 & \textbf{42.58}\\
		Dress     & 14,128 & 4,346  &  869 & 2,825 &       22.93 & 24.90 & 25.56 & 29.33 & 34.34 & \textbf{49.58}\\
		Top      & 7,155 & 2,201     & 440  & 1,431 &       17.45 & 24.83 & 25.26 & 29.14 & 32.94 & \textbf{35.12}\\
		Mini skirt  & 6,571 & 2,021  & 404  & 1,314 &      23.35 & 24.83 & 27.47 & 29.50 & 31.30 & \textbf{32.48}\\
		Hat      & 6,534 & 2,010     & 402   & 1,306 &      15.82 & 13.98 & 20.19 & 25.87 & 33.81 & \textbf{35.12}\\
		Sunglass   & 6,133 & 1,886   & 377 & 1,226 &       11.85 &  7.46 & 11.35 & 11.83 & \textbf{12.26} & 12.16\\
		Bag      & 5,257 & 1,617    & 323  & 1,051 &        23.78 & 27.63 & 27.47 & 25.67 & 25.48 & \textbf{36.82}\\
		Skirt   & 4,453 & 1,370    & 274 & 890 &            19.79 & 25.06 & 22.44 & 24.50 & 24.43 & \textbf{41.75}\\
		Suit     & 3,906 & 1,201      & 240  & 781 &        18.65 & 25.18 & 19.72 & 25.29 & 26.60 & \textbf{42.08}\\
		Shoes      & 3,358 & 1,033    & 206 & 671 &         11.45 & 24.10 & 23.92 & 25.03 & \textbf{27.58} & 26.95\\
		Shorts      & 3,249& 999   &  199 & 649 &           11.15 &  5.99 & 13.90 & 14.84 & \textbf{16.62} & 13.74\\
		Pants     & 2,738 & 842        & 168 & 547 &      17.57 & 22.54 & 25.77 & 29.49 & 28.36 & \textbf{32.13}\\
		Breeches     & 2,044& 628  & 125 & 408 &          23.45 & 22.99 & 25.03 & 28.52 & 28.76 & \textbf{48.28}\\
		High shoots& 2,007 & 617   & 123  & 401 &           12.05 & 13.11 & 14.57 & 15.46 & \textbf{16.04} & 14.94\\
		\hline
		Overall& 85,677 & 26,352  & 5,266 & 17,128 &   18.36 & 21.44 & 23.47 & 25.73 & 28.73 & \textbf{36.63}\\
		\hline\hline
	\end{tabularx}
\end{table*}

\subsection{Comparison With State-of-the-art Approaches}
To verify the effectiveness of the proposed AsymNet, we compare it with the following state-of-the-art approaches:
1) \textbf{AlexNet (AL)} \cite{NIPS12_AlexNet}: the activations of the fully-connected layer fc6 (4,096-d) are used to form the feature representation.
2) \textbf{Deep Search (DS)} \cite{MM14_DS}: it is an attribute-aware fashion-related retrieval system based on convolutional neural network.
3) \textbf{F.T. Similarity (FT)} \cite{ICCV15_Wheretobuy}: category-specific two-layer neural networks are trained to predict whether two features extracted by the AlexNet represent the same product item.
4) \textbf{Contrastive \& Softmax (CS)} \cite{TOG15_CS}: it is based on the Siamese Network, where the traditional contrastive loss function and softmax loss function are used.
5) \textbf{Robust contrastive loss (RC)} \cite{ICMR16_Product}: multi-task fine-tuning is adopted, in which the loss is the combination of contrastive and softmax.
For clothing trajectories in videos, we calculate the average similarity to gain the most similar shopping images.
The cosine similarity is used in all these methods except FT.

The detailed performance comparison is listed in Table \ref{tab:performance}.
AsymNet achieves the highest performance for top-20 retrieval accuracy. It significantly outperforms AlexNet, in which the performance is almost doubled.
The performance of AlexNet \cite{NIPS12_AlexNet} and Deep Search \cite{MM14_DS} is unsatisfactory, which only use the convolutional features to retrieve images and do not learn the underlying similarity,
The performance of two contrastive based methods (CS \cite{TOG15_CS} \& RC \cite{ICMR16_Product}) are slightly better than FT \cite{ICCV15_Wheretobuy}, since contrastive loss has a stronger capability to identify minor differences.
RC has better performance than CS because it exploits the category information of clothing. For some categories having no obvious difference in clothing trajectories, RC performs slightly better than AsymNet.
Overall, our proposed approach shows clearly better performance than these approaches. This is mainly because AsymNet can handle the temporal dynamic variety existing in videos, and it integrates discriminative information of video frames by automatically adjusting the fusion strategy.

\begin{figure}[tb]
	\centering
	\includegraphics[scale=0.33]{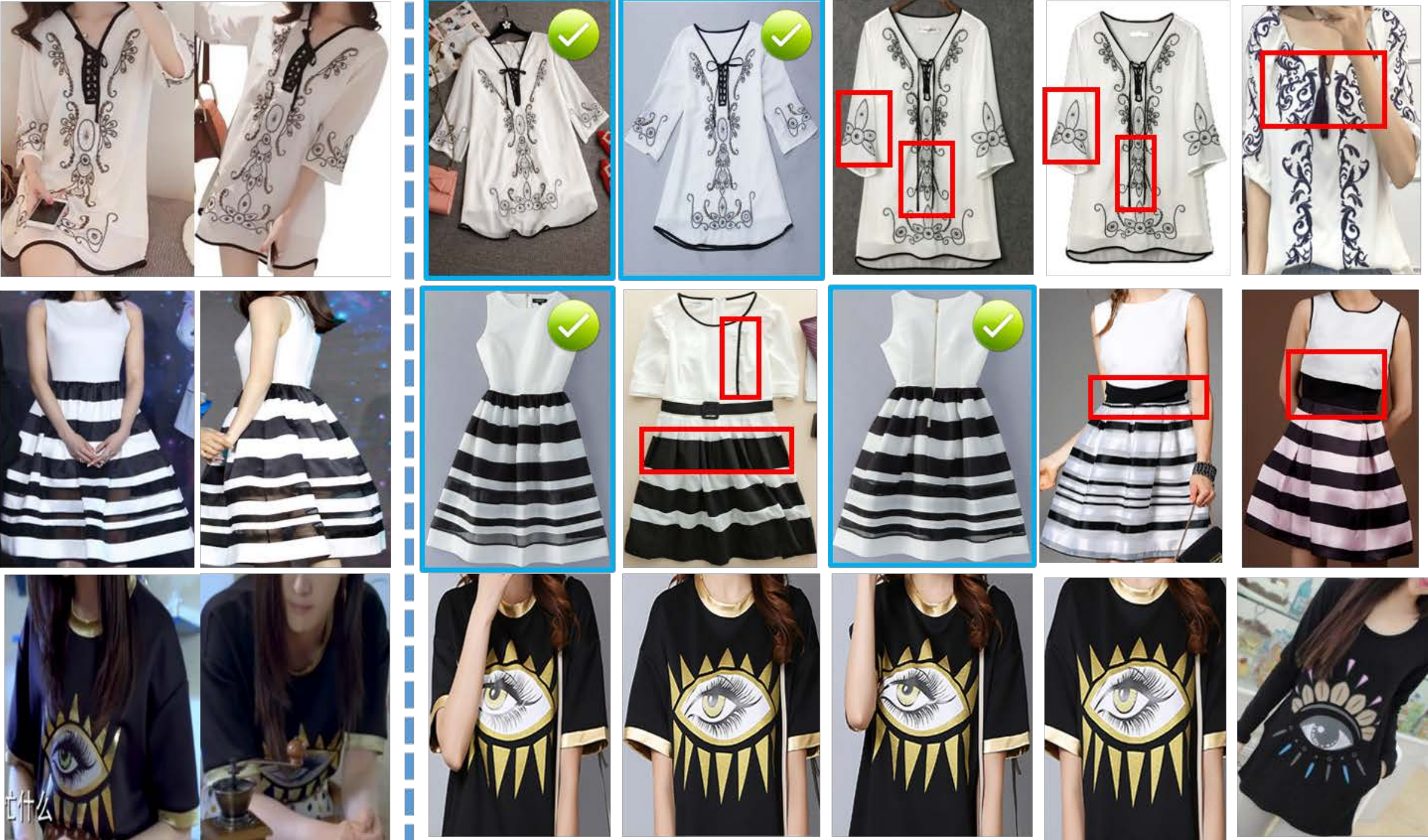}
	\caption{Example with top-5 retrieval results of the proposed AsymNet. The difference in terms of detailed decorative patterns are labelled with red boxs.}
	\label{fig:F4}
	\vspace{-0.2in}
\end{figure}

Three examples with top-5 retrieval results of the proposed AsymNet are illustrated in Fig. \ref{fig:F4}, where the exact matches are marked with green tick. Relatively, it is easier to obtain the visually similar clothes, but it is much challenging to obtain the identical one, especially the query is from videos. For the first two rows, these returned results are visually similar. However, some detailed decorative patterns are different, which are labelled with red boxs. In the last row, although the clothing style is the same, the color is different, so it will not be treated as the correct match.

\subsection{Efficiency}
To investigate the efficiency of the approximate training method, we compare it with traditional training procedure. All these experiments are conducted on a server with 24 Intel(R) Xeon(R) E5-2630 2.30GHz CPU, 64GB RAM and one NVIDIA K20 Tesla Graphic GPUs. In our experiment, \czq{one sample is performed in inference}, the image feature network processes 200 images/sec. The video feature network conducts 0.5 trajectories/sec and the similarity network performs 345 pairs/sec.
The computation can be further pipelined and distributed for large-scale applications. The approximate training only costs 1/25 of the training time of tradition way. Meanwhile, the effectiveness of AsymNet is not influenced with the approximate training method. The training of our AsymNet model only takes around 12 hours to converge.

\section{Conclusion}
In this paper, a novel deep neural network, AsymNet is proposed to exact match clothes in videos to online shops. The challenge of this task lies in the discrepancy existing in cross-domain sources between clothing trajectories in videos and online shopping images, and the strict requirement of exact matching. This work is the first exploration of Video2Shop application. In our future work, we will integrate clothing attributes to further improve the performance.

{\small
	\bibliographystyle{ieee}
	\bibliography{egbib}
}

\end{document}